\documentclass[runningheads]{llncs}
\usepackage{graphicx}
\usepackage{listings}
\usepackage{todonotes}
\usepackage{float}
\usepackage{pdflscape}
\usepackage{afterpage}
\usepackage{capt-of}
\usepackage[ruled,vlined]{algorithm2e}
\usepackage{booktabs}

\begin{document}
\title{Exploring Sequence-to-Sequence Models for SPARQL Pattern Composition}
%
%
\author{Anand Panchbhai\inst{3} \and
Tommaso Soru\inst{1,3} \and
Edgard Marx\inst{1,2,3}
\institute{
AKSW, University of Leipzig, Germany \and
Leipzig University of Applied Sciences, Germany \and 
Liber AI Research, London, UK
}
}
\authorrunning{Panchbhai et al.}
\maketitle              
\begin{abstract}
A booming amount of information is continuously added to the Internet as structured and unstructured data, feeding knowledge bases such as DBpedia and Wikidata with billions of statements describing millions of entities. 
The aim of Question Answering systems is to allow lay users to access such data using natural language without needing to write formal queries. 
However, users often submit questions that are complex and require a certain level of abstraction and reasoning to decompose them into basic graph patterns. 
In this short paper, we explore the use of architectures based on Neural Machine Translation called \textit{Neural SPARQL Machines} to learn pattern compositions. 
We show that sequence-to-sequence models are a viable and promising option to transform long utterances into complex SPARQL queries. 


\end{abstract}
\section{Introduction}

Knowledge graphs have recently become a mainstream method for organising data on the Internet. 
With a great amount of information being added every day, it becomes very important to store them properly and make them accessible to the masses. 
Languages like SPARQL have made querying complex information from these large graphs possible. 
Unfortunately, the knowledge of query languages such as SPARQL is still a barrier that makes it difficult for lay users to access this data readily. 
A number of attempts have been made to increase the accessibility of knowledge graphs. 
Prominent among them is the active application of various paradigms of computer science ranging from heuristic-based parsing methodologies to machine learning. 

Recent advances in Natural Language Processing have shown promising results in the task of machine translation from one language to another. 
Eminent among them are neural architectures for sequence-to-sequence (Seq2Seq) learning~\cite{sutskever2014sequence} in Neural Machine Translation (NMT). 
Considering SPARQL as another language for such a conversion, attempts have been made to convert natural language questions to SPARQL queries by proposing Neural SPARQL Machines~\cite{sorumarx2017,soru2018} (NSpM). 
Rather than using language heuristics, statistical or handcrafted models, NSpM completely rely on the ability of the NMT models.
The architecture majorly comprises of 3 components (i.e., generator, learner, and interpreter), and its modular nature allows the use of different NMT models~\cite{sorumarx2017}. 

This paper tries to answer the problem of Knowledge Graph Question Answering (KGQA) from a fundamental perspective. 
We build upon the NSpM architecture to tackle 2 main research questions:
\begin{enumerate}
    \item[(RQ1)] Can we employ Seq2Seq models to learn SPARQL pattern compositions?
    \item[(RQ2)] Targeting compositionality, what is the best configuration for a NSpM to maximise the translation accuracy?
\end{enumerate}

To address the first question, we start by augmenting the template generation methodology proposed for the NSpM model in~\cite{soru2018}.
This is followed up with an analysis of compositional questions.
This analysis will help gauge the competence of NMT models to learn SPARQL pattern compositions. 
In the later sections, we will lay out the best configuration for NSpM to maximise its translation performance. 
We released all code and data used in the experiments.\footnote{\url{https://github.com/LiberAI/NSpM/wiki/Compositionality}}

\section{Related Work}

Many approaches have tried to tackle the challenge of KGQA. 
In a number of these approaches, an attempt is made to retrieve the answers in the form of triple stores~\cite{unger2012template,shekarpour2015semantic,dubey2016asknow}. 
Questions can be asked in a variety of forms. 
Work has been done specific to simple questions pertaining to facts~\cite{lukovnikov2017neural}, basic graph patterns (BGP)~\cite{shekarpour2015semantic,zhang2016joint} and complex questions~\cite{unger2012template,dubey2016asknow}. 
The advent of QA datasets like QALD~\cite{ngomo20189th,usbeck20188th}, LC-QuAD~\cite{trivedi2017lc}, and DBNQA~\cite{hartmann-marx-soru-2018} has accelerated the use of deep learning based techniques for the purpose of KGQA. 
KQA Pro~\cite{shi2020kqa} is a relatively new dataset that tries to tackle the problems present in previous QA datasets by using recursive templates and crowd-sourced paraphrasing methods, incorporating compositional reasoning capability in complex KGQA. 
Some of the entries of challenges like QALD comprise the state of the art in KGQA. 
The approaches range from a ad-hoc rule-based implementations to end-to-end deep learning pipelines. 

\textit{WDAqua}~\cite{diefenbach2017wdaqua} is a rule-based combinatorial system to engender SPARQL queries form natural-language questions. 
The system is not based on machine learning and does not require any training. 
\textit{ganswer2}~\cite{zou2014natural} uses a semantic query graph-based approach to generate SPARQL queries from natural language questions and redefines the problem of conversion of natural language queries to SPARQL as a sub-graph matching problem. 
It constitutes of 2 stages, where the first stage focuses on understanding the questions and the second stage deals with query evaluation.
The approach won the QALD-9 challenge~\cite{ngomo20189th}.
Other Seq2Seq architectures have been proposed to target structured QA, however to the best of our knowledge, none of them has tackled the SPARQL compositionality problem in neural approaches~\cite{liang2016neural,zhong2017seq2sql,zheng2018question}.


\section{Methodology}

\subsection{Problem statement}

We define the problem of SPARQL compositionality from a machine-learning point of view.
Given two questions $a=$\textit{``When was Barack Obama born?''} and $b=$\textit{``Who was the 44th President of the USA?''}, their composition $a \circ b$ is the composite question \textit{``When was the 44th President of the USA born?''}. 

Let us introduce a set $X$ of questions
\begin{equation}
    X = \left\{ a_1, b_1, a_1 \circ b_1, \ldots, a_{n-1}, b_{n-1}, a_{n-1} \circ b_{n-1}, a_n, b_n \right\}
\end{equation}
mapped to its respective set of queries $Y$.
The basic problem of compositionality is to be able to predict $f(a_n \circ b_n)$ by learning $f : X \rightarrow Y$.
In other words, we expect the learning model to generalise on the seen examples $(X,Y)$ and learn to compose on the unseen example $a_n \circ b_n$.

\subsection{Template generation}

The generator is the first part of the NSpM architecture.
The training data outputted by the generator is heavily dependent on the structure of the input templates. 
Manual curation of templates has been carried out in a number of works~\cite{unger2012template,ngomo20189th,soru2018}. 
Part automated and part manual methods powered by crowd sourcing have been used in various versions of the QALD benchmarks, LC-QuAD, and KQA Pro. 

This work proposes a completely automated way of generating templates, as it follows a bottom-up approach.  
A ranking-based method is proposed to ensure that the templates used are natural and germane to the questions asked by general users. 
The template generation methodology proposed here is for the DBpedia knowledge graph~\cite{lehmann2015dbpedia}.

The first step deals with iterating over all the properties of a given class. 
For each class of the DBpedia ontology, we retrieve metadata such as label, domain, range, and comments.\footnote{The metadata can be fetched from \url{http://mappings.dbpedia.org/server/ontology/classes/}.}
Based on the properties, questions can be constructed. 
For instance, \textit{date of birth} is the property of an individual and is represented by \texttt{dbo:birthDate} in DBpedia.
Information about all entities having a \texttt{dbo:birthDate} value are fetched using SPARQL queries of the form: \texttt{SELECT DISTINCT(?a) WHERE \{ ?a dbo:birthDate [] \}}.  
As individual \texttt{dbr: Barack\_Obama} is one of those entities having a \texttt{dbo:birthDate} value, questions can be framed as \textit{What is the date of birth of Barack Obama?} with corresponding SPARQL query: \texttt{SELECT ?x WHERE \{ dbr:BarackObama dbo:birthDate ?x \}}.
The current question is a very primitive one and has a specific form: \textit{What is the $\langle$Property name$\rangle$ of $\langle$Entity name$\rangle$?}. 
More often than not, users ask more involved questions. 
Given the structural form of the SPARQL language, a bottom up approach was again adopted to build questions of varying types. 

\begin{figure}
    \centering
    \includegraphics[width=0.7\textwidth]{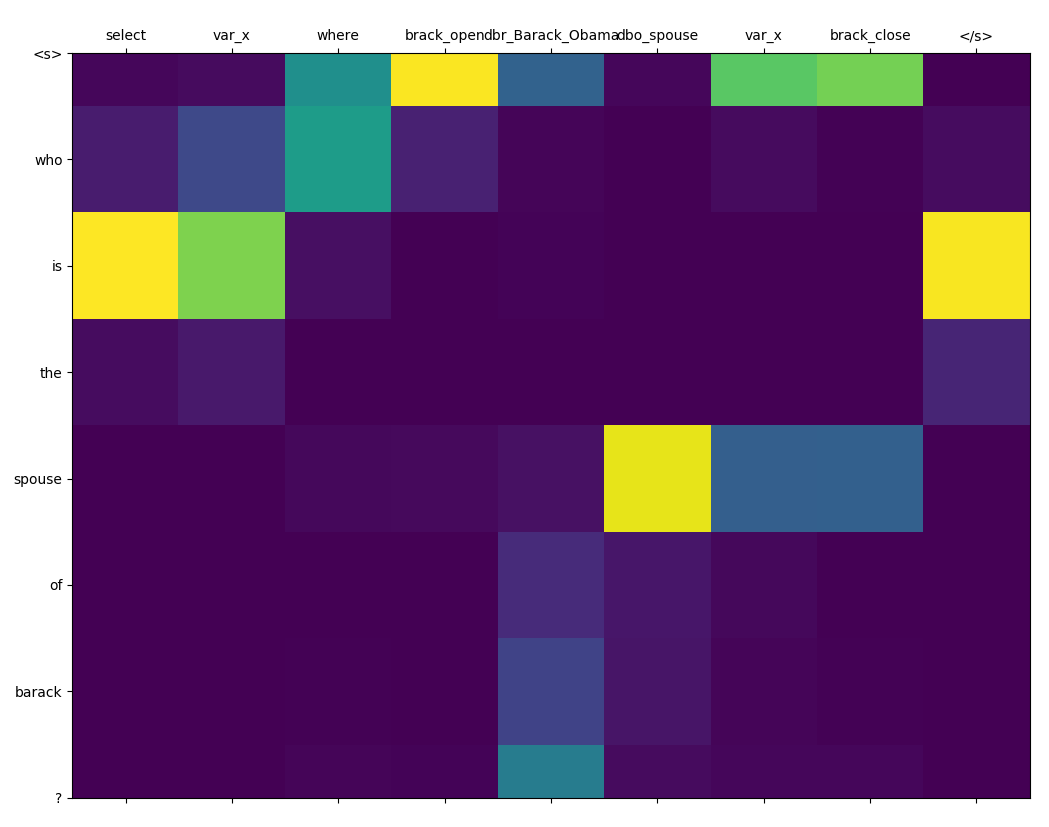}
    \caption{Attention heat-map of a Neural SPARQL Machines at work.
    The x-axis contains the natural language question while the y-axis contains the sequence encoding of the corresponding SPARQL query.  
    }
    \label{fig:heatmap}
\end{figure}

\subsubsection{Types of questions.}
SPARQL supports answering conditional questions using conditional operators. 
Comparative questions are one of the basic types of questions asked as queries. 
Similarly, other functionalities provided by SPARQL such as \texttt{LIMIT} and \texttt{OFFSET} could be used for creating more complex questions pertaining to aggregational queries and questions containing superlatives. 
Questions with boolean answers can be answered using the \texttt{ASK} query form. 
Intuitively, increasingly complex compositional questions can be generated by recursively running the same steps above. 

\subsubsection{Ranking.}

On close inspection, we noticed that a number of templates generated felt unnatural. 
To tackle this issue, we assigned a rank to each template based on a page ranking mechanism on the basis of the hypothesis that the relevance of a template can be determined by the popularity of the corresponding answers. 
The ranking mechanism used here is SubjectiveEye3D\footnote{\url{https://github.com/paulhoule/telepath/wiki/SubjectiveEye3D}}, which is similar to PageRank~\cite{page1999pagerank}. 


The ranking step takes place after the generation and checking of the compatible entities.
For example, in the question \textit{``What is the date of birth of spouse of $<$A$>$?''}, the placeholder $<$A$>$ can be replaced by entity labels (e.g., ``Barack Obama'').
Here, \texttt{dbr:Barack\_Obama} and the required date of birth are the two directly useful entities, whereas the intermediate entity is \texttt{dbr:Michelle\_Obama}. 
A damping factor was introduced as the depth of question increased to take the route of getting the answer into consideration. 
In the previous example, the route is: \texttt{dbr:Barack\_Obama} $\rightarrow$ \texttt{dbr:Michelle\_Obama} $\rightarrow$ \texttt{dbo:birthDate}. 
0.85 was selected as a damping factor empirically, following the probability arrived in a similar research done in \cite{brin1998anatomy}. 
The accumulated scores for a given templates were averaged to get the final rank.
The decision to consider a given template is based on a threshold mechanism.
The threshold should be decided class-wise and a single general threshold should not be used for all classes, since the pages related to certain classes (e.g., eukaryotes) are viewed less than pages related to others (e.g., celebrities). 
The relative number of views within eukaryotes-related pages is however useful for the ranking of templates. 

\section{Results and discussion}

We employed a parameter search where we varied various parameters of the NSpM architecture.
The results obtained are shown in Table~\ref{tab:results}. 
Due to the sheer size of DBpedia, we selected the subset of entities belonging to the eukaryotes class to carry out our compositionality experiments. 
Eukaryotes is a large class of DBpedia with 302,686 entities and 2,043 relations.\footnote{Retrieved on 19/10/2020 from \url{https://dbpedia.org/sparql}.}
169 unique templates were generated by the automatic template generation pipeline, these templates yielded 21,637 unique natural language question-query pairs.
The best results for the given dataset were obtained for the configuration of NMT with 2 layers, 128 units, 0.7 dropout, scaled Luong attention and with pre-trained embeddings. 
Ensuring that the entities were present in the training set for a predetermined number of times also helped boost the translation performance, as previously found in~\cite{sorumarx2017}. 
Use of attention in the NMT architecture helped the model in it's translation capabilities as is evident from the results present in table \ref{tab:results}. 
Figure \ref{fig:heatmap} depicts the attention weights for a translation of a natural-language question into a sequence encoding a SPARQL query in the form of a heat-map.

\begin{table}
\caption{\textbf{Parameter search results.} Each model configuration was run for at least 40,000 epochs. All options of the type column have the following meaning: (a) random splitting into 80\% training, 10\% validation, and 10\% test sets; (b) ensuring same vocabulary of training and test sets; (c) frequency threshold ensured that the training set had a predetermined number of iterations of a given word to give the model ample opportunity to learn the word to be tested; (d) the restrictions of experiment \#1.2 were followed. Best BLEU and accuracy are given in percentage.
}

\begin{tabular}{p{0.8cm} c c c c c c r r}
\toprule
{\textbf{\#}} & \textbf{Type} & \textbf{\#Layers} & \textbf{\#Units} & \textbf{Dropout} & \textbf{Attention} & \textbf{Emb.} & \textbf{BestBLEU} & \textbf{BestAcc.} \\ \midrule
\multicolumn{9}{c}{Attention or no Attention} \\ \midrule
1.1 & a & 2 & 128 & 0.2 & No & No & \textbf{97.7} & \textbf{89.8} \\ \midrule
1.2 & a & 2 & 128 & 0.2 & Scaled Luong & No & 97.5 & 88.0 \\ \midrule
1.3 & b,c,d & 2 & 128 & 0.2 & No & No & 66.4 & 5.7 \\ \midrule
1.4 & b,c,d & 2 & 128 & 0.2 & Scaled Luong & No & 85.2 & 34.3 \\ \midrule
\multicolumn{9}{c}{Dropout} \\ \midrule
2.1 & b,c,d & 2 & 128 & 0.05 & Luong & No & 58.0 & 0.0 \\ \midrule
2.2 & b,c,d & 2 & 128 & 0.5 & Luong & No & 86.0 & 40.9 \\ \midrule
2.3 & b,c,d & 2 & 128 & 0.7 & Luong & No & 85.0 & 45.1 \\ \midrule
2.4 & b,c,d & 2 & 128 & 0.9 & Luong & No & 59.6 & 2.3 \\ \midrule
\multicolumn{9}{c}{Attention type} \\ \midrule
3.1 & b,c,d & 2 & 128 & 0.5 & Luong & No & 76.7 & 9.1 \\ \midrule
3.2 & b,c,d & 2 & 128 & 0.5 & Bahdanau & No & 62.5 & 0.0 \\ \midrule
3.3 & b,c,d & 2 & 128 & 0.5 & Scaled Luong & No & 86.0 & 40.9 \\ \midrule
\multicolumn{9}{c}{ Number of Units} \\ \midrule
4.1 & b,c,d & 2 & 256 & 0.5 & Scaled Luong & No & 82.9 & 25.0 \\ \midrule
4.2 & b,c,d & 2 & 512 & 0.5 & Scaled Luong & No & 55.8 & 0.0 \\ \midrule
\multicolumn{9}{c}{Number of Layers} \\ \midrule
5.1 & b,c,d & 1 & 128 & 0.7 & Scaled Luong & No & 1.0 & 0.0 \\ \midrule
5.2 & b,c,d & 3 & 128 & 0.5 & Scaled Luong & No & 58.7 & 0.0 \\ \midrule
5.3 & b,c,d & 2 & 128 & 0.5 & Scaled Luong & No & 86.0 & 40.9 \\ \midrule
5.4 & b,c,d & 4 & 128 & 0.5 & Scaled Luong & No & 63.0 & 0.0 \\ \midrule
\multicolumn{9}{c}{Pre-trained embeddings} \\ \midrule
6.1 & b,c,d & 2 & 128 & 0.7 & Scaled Luong & Yes & 93.0 & 63.0 \\ \bottomrule
\end{tabular}
\label{tab:results}
\end{table}

For answering the question related to compositionality and the ability of the Seq2Seq NMT model, experiment \#1.1 was carried out. 
As previously introduced, by compositionality here we mean that $a$ and $b$ were introduced in the training phase, we test for $a \circ b$ or $b \circ a$. 
$a$ and $b$ represent the properties or entities. 
On randomly splitting the dataset thus generated into 80\% train, 10\% validation and 10\% test, we were able to achieve 97.69\% BLEU score and 89.75\% accuracy on 40,000 iterations of training. 
Perfect accuracy was not achieved due to (1) entity mismatch and (2) certain instances where the training set did not contain the entities that were present in the test set.
These issues can be tackled by increasing the number of examples per entity in the training set, however it is unrealistic to expect an algorithm to disambiguate all entities, as they may be challenging even for humans.


Though the model produced favourable results, a more challenging task was created to assess the ability of Seq2Seq NMT model towards learning pattern compositions. 
A special pair of training and test set (experiment \#1.2) was created while putting a restriction on the entities present in the training set. 
In this regard, the training set could contain templates $a, b, a \circ c, d \circ b$, whereas the test set contained $a \circ b, a, b, c, d$. 
A property once introduced in given depth was not introduced there again in the training set; instead, it was added to the test set.
These results are represented in the last row in Table~\ref{tab:results}.
As can be seen, we obtained a BLEU score of 93\% and accuracy of 63\%. 
The results thus obtained tell us few very important things about the ability of the NMT model, as even with less amount of data per entity the NMT model was able to learn the structure of SPARQL query that needed to be generated. 
This shows the ability of Seq2Seq to adapt to complex language structure whenever necessary. 
On analysing the result, it was discovered that the low accuracy and high BLEU was again due to the entity mismatch that took place despite the model predicted the right structure of the queries. 
The model in the previous experiment was able to produce high accuracy merely because it had more opportunity to learn the given word and structure when compared to experiment 1.2. 

While conducting these experiments, we were able to gauge the importance of various parameters which are an essential part of template generation and the NSpM model. 
The study carried out here and the results stated  are limited to the particular ontology class of Eukaryotes. 
Although the generated system is not potent to answer questions beyond this class, later studies can train on more number of classes to address this problem. 
The ability of the model to handle more complex questions with varying template structure also needs to be explored. 

\section{Conclusion}

This study suggested a way to generate templates automatically for the NSpM pipeline. 
An optimal configuration was also suggested based on the experiments conducted as part of the study.
The results suggest that Seq2Seq NMT model holds the potential to learn pattern compositions.
We plan on making the generated templates sound more human by integrating NLP paraphrasers and pre-trained language models such as GPT~\cite{brown2020language} and BERT~\cite{devlin2018bert}.

This work was partly carried out at the DBpedia Association and supported by Google through the Google Summer of Code 2019 programme.

\bibliographystyle{splncs04}
\bibliography{refs}

\end{document}